# Soft Robots for Extreme Environments: Removing Electronic Control*

Stephen T. Mahon, *Member, IEEE*, Anthony Buchoux, Mohammed E. Sayed, Lijun Teng, and Adam A. Stokes, *Member, IEEE*

*Abstract*—The ignition of flammable liquids and gases in offshore oil and gas environments is a major risk and can cause loss of life, serious injury, and significant damage to infrastructure. Power supplies that are used to provide regulated voltages to drive motors, relays, and power electronic controls can produce heat and cause sparks. As a result, the European Union requires ATEX certification on electrical equipment to ensure safety in such extreme environments. Implementing designs that meet this standard is time-consuming and adds to the cost of operations. Soft robots are often made with soft materials and can be actuated pneumatically, without electronics, making these systems inherently compliant with this directive. In this paper, we aim to increase the capability of new soft robotic systems moving from a one-to-one control-actuator architecture and implementing an electronics-free control system. We have developed a robot that demonstrates locomotion and gripping using three-pneumatic lines: a vacuum power line, a control input, and a clock line. We have followed the design principles of digital electronics and demonstrated an integrated fluidic circuit with eleven, fully integrated fluidic switches and six actuators. We have realized the basic building blocks of logical operation into combinational logic and memory using our fluidic switches to create a two-state automata machine. This system expands on the state of the art increasing the complexity over existing soft systems with integrated control.

## I. Introduction

A major risk in offshore oil and gas environments is the ignition of a flammable liquid or gas; electrical sparks, static electricity, and friction ignition have been known to cause ignition. The European Union requires that member states follow the *appareils destinés à être utilisés en amosphères explosives* (ATEX) directive which concerns the use of equipment and protective systems in potentially explosive atmospheres. ATEX certification is an involved process requiring a notified body to certify, test, and evaluate the design of the equipment.

Soft robots are often made with soft materials and are actuated using pneumatics, making these systems inherently ATEX compliant and removes the need for a notary. These systems can operate in hostile or poorly accessible environments [1], [2] and can manipulate objects of various size and shape using soft robotic grippers [3]–[5]. To provide technological innovations to support increasing energy demand of our increasing population while maintaining safe and cost-effective production of oil we must increase the

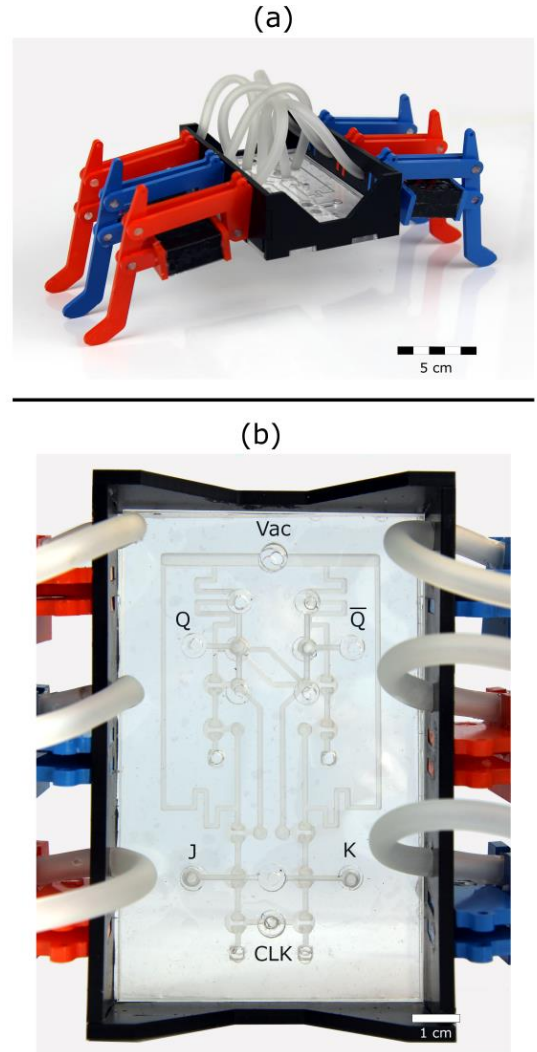

**Figure 1.** An integrated and logically controlled soft robot. (a) The soft robot controls six actuators connected to the legs, colored orange and blue. The controller circuit is designed to engage the orange actuators and the blue actuator sequentially. (b) The fluidic architecture shows a JK flip-flop. The circuit has three inputs including a vacuum power line, a clock line, and a control input, with six outputs to vacuum actuators from Q and $\bar{Q}$.

*This work was supported by the Engineering and Physical Sciences Research Council (EPSRC) via the Off-Shore Robotics for Certification of Assets (ORCA) Hub (EP/R026173/1).

Stephen T. Mahon, Mohammed E. Sayed, Lijun Ting, and Adam A. Stokes are with the School of Engineering, Institute for Micro and Nano Systems, The University of Edinburgh, The King's Buildings, Edinburgh EH9 3LJ, UK (phone: +44-131-650-5611; email: S.Mahon@ed.ac.uk, M.Mohammed@ed.ac.uk, L.Ting@ed.ac.uk, Adam.Stokes@ed.ac.uk).

Anthony Buchoux is with the School of Engineering, Institute for Multiscale Thermofluids The University of Edinburgh, The King's Buildings, Edinburgh EH9 3LJ, UK (email: A.Buchoux@ed.ac.uk).

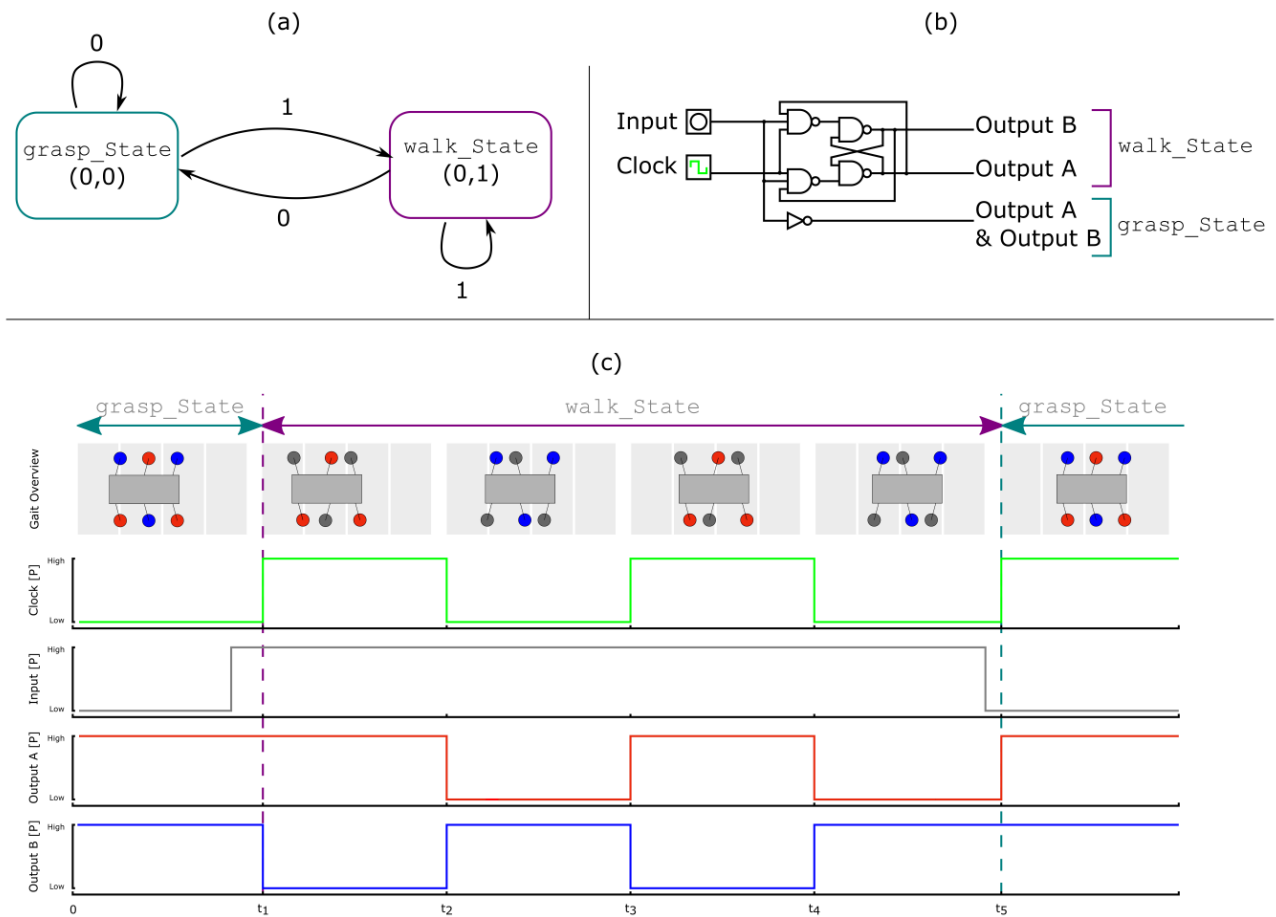

**Figure 2.** The behavior of the soft robotic system. (a) The state diagram is represented as the two states; a grasp state and a walk state. A high input from the idle grasp state moves the system into the walk state. A high input in the walk state keeps the robot in that state, only changing back to a grasp state when the input changes to low. (b) The logical diagram is constructed from the state machine. The diagrams consist of a T flip-flop and a NOT gate. The outputs of the flip-flop alternate when the input is high and the clock switches between the rising edge and falling edge, giving the desired behavior of the walk state. When the input is low the outputs of the flip-flop remain in the last position and grasp state is engaged from the NOT gate. (c) The timing diagram shows the expected behavior of the soft robotic system. The walking state of the robot is inspired by the alternative tripodal gait of an insect.

capabilities of our robots into new areas of exploration, surveillance, and automation.

In recent work, we have shown that as we increase the capabilities of soft robots, the number of controller outputs to actuators increases linearly [6]. This one-to-one mapping architecture physically constrains the system and leads to practical limits in control. Hybrid-soft robots follow the same architecture with the capability of the robot scaling with the number of outputs for control hardware.

In this paper, we aim to increase the capability of new soft robotic systems moving from a one-to-one control-actuator architecture and implementing an electronics-free control system. We have developed a robot that demonstrates locomotion and gripping using three-pneumatic lines: vacuum, clock, and control. This architecture enables a three-to-N mapping for autonomy in movement and manipulation. Increasing the complexity of our electronic-free control system will increase our ability to implement desired motions and behaviors.

We have implemented this electronic-free architecture using fluidic switches, or transistors, consisting of a thin layer of polydimethylsiloxane (PDMS) sandwiched between two sheets of laser cut acrylic. We arranged the fluidic switch to control a gate between the vacuum source and an atmospheric vent. We can use many fluidic transistors from a single vacuum input to remove the requirements for electronic or electro-pneumatic control and minimize the number of outputs from control hardware.

*A. Literature*

The recent review by Shukla and Karki [7] presented a technical overview of robotics used in the oil and gas industry. Robotic assistance and automation are key for the safe and cost-effective production of oil for the rapidly increasing world population. Operation in offshore oil and gas environments requires ATEX certification due to the risk of ignition of flammable liquids and gases. The cost of implementing ATEX certification is significant to add to the cost of operation in the harsh and inaccessible environment. Soft robotics offer some advantages in the cost of production. The soft materials used to fabricate the robotics are often inexpensive and systems can be rapidly prototyped using new manufacturing techniques.

A robotic system is a combination of hardware and control. The prevalent paradigm for the control architecture in soft robotics is a one-to-one mapping of controller outputs to actuators. Recently, we observed that stacking a functional blocks results in systems that are increasingly capable of a diverse range of complex motions [6]. We are beginning to reach practical limits in control due to size restrictions of pneumatic lines and pressure limitations across large pneumatic networks. As we add more functional blocks, we will hit a limit with the number of parallel control lines.

The Arthrobot created complex motion by actuating several of made of arachnid-inspired joints [8]. In their publication, the authors demonstrated increased complex motion with each additional actuator. Tolley et al. [1] developed an untethered soft robot with a battery-driven compressor to provide pressured gas, valves, and a microcontroller. Their soft robot demonstrated resilience to extreme environmental conditions and locomotion over uneven and slippery surfaces.

Integrating soft robotic components with other types of robotic systems offers advantages control and hardware and can create a system greater than the sum of its parts. For example, Stokes et al. [9] and McKenzie et al. [10] developed hybrid robots for grasping and manipulation. A hard robotic component provided definite and fast positioning of the end effector while a soft gripper produced an enveloping grip that was tolerant to positioning error and object irregularity.

Recently, Wehner et al. [11] have shown a fully integrated design and fabrication strategy for an entirely soft autonomous robot. This untethered, pneumatic robot uses a monopropellant decomposition regulated to an actuator without using electronics. This system-level architecture is represented as an electrical analogy: check valves as diodes, fuel tanks as supply capacitors, reaction chambers as amplifiers, actuators as capacitors, vent orifices as pull-down resistors. The controller of Octobot was based on work by Mosadegh et al. [12]. Microfluidic logic autonomously regulated fluid flow for the control system and routed the power source. Mosadegh provided the flow switching acting as a clocking function. The networks of fluidic gates spontaneously generated cascading and oscillatory flow output using only a constant flow of fluids as the device input.

There have been several groups developing fluidic valves, logic circuits, and fluidic processors [13]–[17]. These designs are based on Quake-type valves; microfluidic systems containing switching valves and pumps entirely out of elastomer [18]. An elastomeric membrane block or allows flow through a channel depending on an applied pressure to a gate. Thorsen et al. later developed large-scale integrated microfluidic chips that contained hundreds of addressable chambers accessed using thousands of Quake valves [19]. The microfluidic networks drew analogies of a comparator array with fluidic memory.

Duncan et al. used precision machining techniques to build a variety of digital logic circuits using fluidics [20]. The used a computer numerically controlled (CNC) machine to make normally closed pneumatic membrane valves from Grover and Jensen [13], [21] and demonstrated an operating T flip-flop. Their scaling strategies increased the density of valves to build a 12-bit asynchronous counter circuit that required only a single vacuum connection as a supply power to 108 gates.

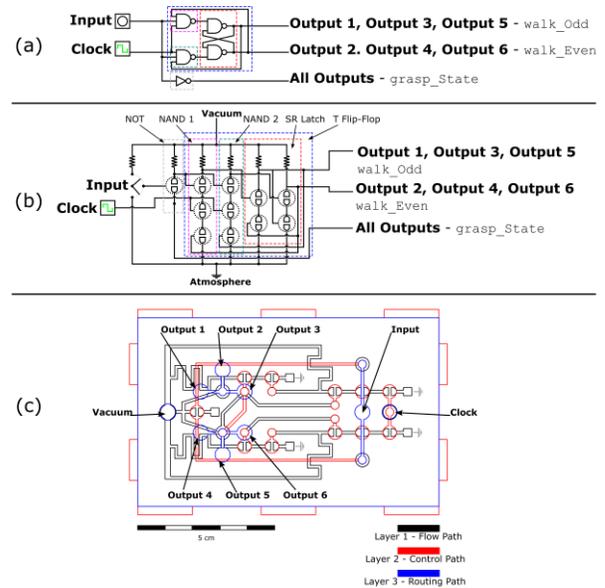

**Figure 4.** Design of the fluidic circuit. (a) The logical diagram as presented in Fig. 2 (b). The T flip-flop consists of two three-input NAND gates and an SR Latch. (b) The fluidic transistor-transistor logic is constructed from the architecture in Fig. 3. (c) The circuit layout includes 3 layers, the flow path, the control path, and the routing layer. The control path turns on and off the fluidic transistors, while the routing path directs the outputs for interfacing with the actuators. A PDMS membrane is sandwiched between the layers with vias for flow through the layers.

We use similar systems to develop a fluidic circuit with combinational logic from a standard fluidic switch as a test of the capabilities of fluidic logic and used this fluidic circuit to control a soft robot. We show an alternating tripodal gait of soft robotic legs demonstrated using a T flip-flop which toggles the output high and low with a corresponding clock signal. Furthermore, we demonstrate a state change within a soft robot without the use of electropneumatic valves. The illustrated logical ideologies will allow extension of the technology to even more complex behaviors for the future of the offshore oil and gas industry, such as manipulation of critical equipment, surveillance providing prevention and cleaning of oil spills, and safety and productivity with remote operation.

*B. Experimental design*

This paper describes an architecture of fluidic switches for the programmable control of soft robots. Grover and Jensen [13], [21] described the use of fluidic switches for the manipulation of very small amounts of fluids for biochemical application with channels of the order of 100 µm. Here we designed our switches and circuits for the capacities of actuators, several orders of magnitude greater than previously reported.

We used a systems engineering approach for stacking and hierarchy for the design of the soft robot as we have described previously [6]. We first defined the behavior of our system and identifying the requirement for the task. This behavior was fully described and decomposed into a set of functions and we described a functional block with the minimum behavior necessary.

We demonstrate completely electronic free control of a soft robot which can perform task-oriented work by designing the fluidic circuit around a desired behavior: locomotion and grasping. Fig. 1 provides an overview of the system. The design of the robot is bioinspired to imitate the locomotion of a hexapod.

### 1) Bioinspired design

Biologically inspired approaches are widely adopted in robotics. Robots that use bioinspired solutions show capabilities for adaptive and flexible interactions with unpredictable environments [22]–[25]. In harsh terrains such as offshore oil rigs, an alternating tripodal gait can offer improved stability as there are always three legs on the ground.

We take advantage of the soft actuators to create a simple design that demonstrates a two-state machine. The restoration force of a soft muscle actuator can be used in place of springs commonly found in traditional robotics. We use only a single actuator per leg module for a simplified design.

### 2) State machine

A typical interaction of a robot is locomotion and manipulation. We chose a walking state based on the alternating tripod gait inspired by an insect. We define the alternating state as `walk_Even` and `walk_Odd`. Analyzing the gait reveals that `walk_Odd` is the negation of the state `walk_Even`. The grasping state can be represented by engaging the appendage of the soft robot synchronously.

We generate the state machine based on these two states in Fig 2 (a). The system is initiated in a `grasp_State`. This state is ideal to keep the soft robot idle as a low input keeps the robot stationary. A high input changes the state from `grasp_State` to `walk_State`. A high input in `walk_State` maintains this state.

Asynchronous state machines require combinational logic and a clock function. The `walk_State` is the oscillation between `walk_Odd` and `walk_Even` states. We can see the logical diagram in Fig. 2 (b). A high input on the flip-flop toggles the output with the rising and falling edge of the clock. The outputs of the T flip-flop, Q and $\overline{Q}$, switch back and forth to give the desired behavior. When the input to the T flip-flop is low, the signal is routed through a NOT gate the activate all the actuators, resulting in the `grasp_State`. The timing diagram in Fig. 2 (c) shows the expected behavior of the robot.

### 3) Fluidic architecture

Previously reported fluidic transistors have been used for the manipulation of fluids for chemical and biochemical application[13]–[17]. We are implementing this architecture for the control of a soft robot rather than for very small amounts of fluids. Here we designed our fluidic transistors and circuits for faster flow rates and larger capacities used for actuators, several orders of magnitude greater than biochemical application.

The fluidic transistors are simple to fabricate; we use a laser cutter to raster channel in multiple layers of acrylic sheets and thin PDMS layers act as a membrane to block or allow flow through a chamber. The PDMS membrane is sandwiched between two layers of acrylic effectively sealing the channels.

We have implemented a normally closed gate. That is, there is no flow through the gate unless a vacuum is applied to the membrane, opening the gate. Fig 3. (a) illustrates how this normally closed gate can be used as the negation operation. When a vacuum is applied at A, the gate is opened, and the vacuum pulls from atmospheric pressure or the active ground. The gate is closed when there is no vacuum at A and instead, the vacuum pulls from the next path of least resistance, giving the output of the operation as the negation of the value at A.

The logical diagram in Fig. 2 (b) is required to enable the desired behavior of the soft robot. This is a two-state machine and requires 1 bit of memory. There are two components on the diagram: a fluidic NOT gate and a fluidic NAND gate that takes a varying number of inputs. The memory of the machine, the T flip-flop, is the combination and arrangements of NAND gates in the diagram.

All Boolean functions can be constructed from the primitive NOT operation. For instance, the fluidic NAND gate is built from two fluidic NOT primitives, and the gated SR latch is made from two cross-coupled NAND gates. The SR latch is the simplest bi-stable device that enables memory in a system. Fig. 3 (b) and (c) shows these more complex operations built from the fluidic NOT primitive.

## II. RESULTS AND DISCUSSION

The fluidic circuit under investigation is outlined in Fig. 4 (c) which illustrates a T flip-flop circuit fabricated etched on acrylic using a laser cutter. The finished circuit can be seen in Fig. 1 (b). The alternating tripodal gait robot can be seen in Fig. 5. The fluidic circuit with the soft robot demonstrates a two-state machine. A high input and an alternating clock cycle demonstrate a walking motion as outlined with the behavior of the system. When the input is low, the flip-flop remains in the last state and the low input travels through a NOT gate to engage all actuators for a grasp action.

## III. PREPARATION OF MATERIALS

### A. Fluidic Switches

The fluidic circuit was fabricated on a CNC milling machine. We designed the circuit on the 3D CAD design software Fusion 360 and exported the 3D designs into STL

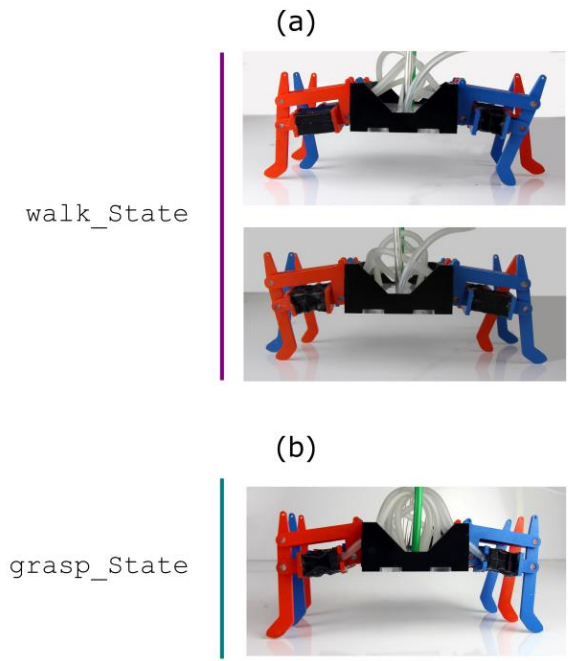

**Figure 5.** Soft robot actuation from a front view. (a) The soft robot autonomously alternating between even and odd actuation states when the input is high regulated through the fluidic circuit outlined in Fig.4. (b) A low input changes the state to the grasp, with all actuators engaging.

files for milling. The circuit was designed with 1 mm channels.

The mechanical properties of the membrane affect the capacitance of the gates which controls the timing of the fluidic circuit. We experimented with the capacitances of the circuit by varying the thickness, between 200 μm and 2 mm, curing time, from 1 hour to 24 hours, and curing temperature, between 20 °C and 100°C, of the PDMS. We opted for a 1 mm layer of PDMS Sylgard® 184 Silicone Elastomer (Dow Corning®). The PDMS is mixed as a ratio of 10:1 and cured for 1.5 hours at 60-80 °C.

### B. Soft robot assembly

The actuators in this paper are based on the work by Yang at al. [26]. These vacuum-actuated muscle-inspired pneumatic structures (VAMPs) generate linear motion under pressure utilizing a buckling structure. The vacuum actuators are to operate in harsh environments; the actuators are still functional even if punctured.

The VAMPs restore to a resting position when returned to ambient pressure. We experimented with different materials PDMS Sylgard® 184 Silicone Elastomer (Dow Corning®), Ecoflex-0030 (Smooth-on, Inc.), Ecoflex-0050 (Smooth-on, Inc.), and Dragon Skin® 30 (Smooth-on Inc.). We opted for Dragon Skin® 30 (Smooth-on Inc.) as it provided the best restoration from the applied vacuum. We 3D printed a negative mold in two halves for the internal structure of the VAMPs. We followed the literature closely scaling the size of the actuator by half while maintaining the same ratio of chamber size to wall thickness. We cured the mixture at room temperature for 3 hours and bonded the halves using the same material at 60 °C for 15 minutes.

The leg module of the soft robot is a rigid link made from acrylic with one degree of freedom, a design choice to keep the circuit design simple. The VAMPs restore to a resting position at ambient pressure removing any requirement for a restoration spring on the leg module. We laser cut the module from 3 mm acrylic with a 10-degree angle to the chassis of the robot. We cut the chassis of the robot from 3 mm acrylic on a laser cutter. the control layer of the fluidic circuit is integrated into the base of the chassis.

## IV. Discussion

Most soft robots use a controller located outside of the system. The Arthrobot was designed this way to observe an emergent behavior. When the control system is offloaded external to the robot the one-to-one mapping of control hardware actuators places physical maximum constraints to the soft system. Untethered robots use a microcontroller to direct flow to the actuators. This method increases the capabilities of the soft robot to be used in new areas of research and exploit the advantages of soft systems.

The Octobot is of extreme importance in soft robotics; the robot combines control and flow-path, intersecting robotics and fluidic controls. The system-level architecture was represented as an electrical analogy with check valves, reaction chambers, actuators and vent orifices as diodes, capacitors, amplifiers, and pull-down resistors. The behavior we have demonstrated with our soft robot is like the Octobot; two groups of actuators that oscillate. The observed behavior of the Octobot was implemented using a monopropellant decomposition regulated to actuators through an embedded microfluidic logic controller. We have expanded on the work from the Octobot to demonstrate a system for task-orientated work with the oscillation of the actuators regulated using a clock signal rather than the capacitances of the system.

We have realized the basic building blocks of logical operation into combinational logic and memory using fluidic switches inspired by Grover et al. [14] to create a two-state automata machine: a walking state, and a grabbing state. This expands on the work from the Octobot which included two gates. We have increased the complexity of our system by an order of magnitude compared to the Octobot.

We believe that fluidic switches are a stepping stone to creating more complex soft robotic systems. Logical operations combined with switching operations are the cornerstone of modern electronics. The architecture used in our system has the potential to create much more complex systems than currently exists.

## V. Conclusion

The move from analog electronics to digital systems enabled, in part, a digital revolution. Here we have followed the design principles of digital electronics and demonstrated an integrated fluidic circuit with eleven, fully integrated fluidic switches. We have made a two-state automata machine with one bit of memory. The flip-flop is a reusable module and can expand the capabilities into more states and towards autonomy in soft robots. We believe that with continuing research into fluidic circuits containing thousands of cascading flip-flops will enable soft robotic systems to perform much more complex functions and behaviors that we see today.

Soft robots designed in this way may be an ideal candidate for use in offshore oil and gas environments due to the ATEX compliance, low cost, and resilience to extreme environments. Typical behaviors needed for such environments are manipulation of critical equipment and surveillance for prevention and cleaning of oil spills. Current solutions in robotics are not adequate for these operations without ATEX certification by a notified body. The certification can be provided but the cost of provision maybe much greater by several orders of magnitude than soft systems.

We have noted that the one-to-one mapping of control hardware to actuator places physical constraints on the maximum size of the system, limiting the capability of the system. Our fluidic architecture moves towards fewer outputs from control hardware while increasing the number of functional capabilities of the system and moves a step closer to autonomy in soft robotics.

ACKNOWLEDGMENT

The authors thank the members of The Stokes Research Group at The University of Edinburgh.